\documentclass{ecai}
\usepackage{graphicx}
\usepackage{latexsym}
\usepackage{amsmath}
\usepackage{float}
\usepackage{multirow}
\usepackage{setspace}
\usepackage{subfigure}
\usepackage{booktabs}
\usepackage{xspace}

\newcommand{\vocalfolds}{\textit{vocal folds}\xspace}
\newcommand{\void}{\textit{void}\xspace}
\newcommand{\othertissue}{\textit{other tissue}\xspace}
\newcommand{\glottalspace}{\textit{glottal space}\xspace}
\newcommand{\pathology}{\textit{pathology}\xspace}
\newcommand{\surgicaltool}{\textit{surgical tool}\xspace}
\newcommand{\intubation}{\textit{intubation}\xspace}
\newcommand{\colorVocalFolds}{light green\xspace}
\newcommand{\colorPathology}{purple\xspace}
\newcommand{\colorOtherTissue}{green\xspace}
\newcommand{\colorGlottalSpace}{blue\xspace}
\newcommand{\colorSurgicalTool}{red\xspace}
\newcommand{\colorIntubation}{yellow\xspace}

\begin{document}

\begin{frontmatter}

\title{Data Augmentation: a Combined Inductive-Deductive Approach featuring Answer Set Programming}

\author[A]{\fnms{Pierangela}~\snm{Bruno}\orcid{0000-0002-0832-0151}\thanks{Corresponding Author. Email: pierangela.bruno@unical.it.}}

\author[A]{\fnms{Francesco}~\snm{Calimeri}\orcid{0000-0002-0866-0834}}

\author[A]{\fnms{Cinzia}~\snm{Marte}\orcid{0000-0003-3920-8186}}

\author[A]{\fnms{Simona}~\snm{Perri}\orcid{0000-0002-8036-5709}}

 \address[A]{Department of Mathematics and Computer Science, University of Calabria, Rende, Italy}
\address[B]{Short Affiliation of Second Author and Third Author}

\begin{abstract}
Although the availability of a large amount of data is usually given for granted, there are relevant scenarios where this is not the case; for instance, in the biomedical/healthcare domain, some applications require to build huge datasets of proper images, but the acquisition of such images is often hard for different reasons (e.g., accessibility, costs, pathology-related variability), thus causing limited and usually imbalanced datasets.
Hence, the need for synthesizing photo-realistic images via advanced Data Augmentation techniques is crucial.
In this paper we propose a hybrid inductive-deductive approach to the problem; in particular, starting from a limited set of real labeled images, the proposed framework makes use of logic programs for declaratively specifying the structure of new images, that is guaranteed to comply with both a set of constraints coming from the domain knowledge and some specific desiderata.
The resulting labeled images undergo a dedicated process based on Deep Learning in charge of creating photo-realistic images that comply with the generated label.
\end{abstract}

\end{frontmatter}

\section{Introduction}\label{sec:intro}
In recent years, applications of Deep Learning (DL) gained a lot of popularity, due to the impressive results in many areas such as image processing, pattern, and object recognition~\cite{wu2019review}. 
However, such methodologies are based on models that have to be trained over some background knowledge represented in proper data; in order to obtain accurate training, large collections of data are typically needed. 
In some domains, obtaining a fair amount of training data is not easy; for instance, this is particularly challenging in the biomedical domain, due to acquisition accessibility, costs, manual annotation effort, data availability, and imbalance. 
To overcome this problem, data augmentation techniques have been widely studied in order to enrich (and improve to several extents) poor datasets.
To this respect, generative models such as Generative Adversarial Networks (GAN) have been proposed to create synthetic but realistic images, showing a great deal of potential.
Nevertheless, such approaches present also some drawbacks and limitations. 
For example, their training can be unstable and slow~\cite{durgadevi2021generative}; furthermore, in general, to guide feature extraction and image generation one has to rely on proper composition and adaptation of the dataset used for training.
This makes difficult to take advantage from available knowledge and express desiderata on the way data will be generated. 
However, such knowledge can be of great help in avoiding the generation of wrong images, reducing generation times, improving the overall quality of the results (e.g., by ensuring the generation of reliable images that respect some given directions/desiderata, preserving the nature of the data, protecting relevant features). 
The use of declarative approaches can help at expressing constraints emerging from the background knowledge along with specific desired features; for this reason, the design of hybrid solutions featuring inductive and deductive techniques should be explored.

In this work, we make a first step in this direction and  propose the use of Answer Set Programming (ASP) for incorporating express knowledge that guides the automatic generation of realistic images.
In particular, we consider image generation in the biomedical domain, with a special focus on a Laryngeal Endoscopic Dataset~\cite{laves2019dataset}.
The idea is to start from a (even limited) set of available photo-realistic images. 
First, a number of labeled images are produced; then, relevant elements are identified that make images differ one from another, and ASP is used for producing brand-new labeled images by describing how such elements can appear according to a given background (medical) knowledge. 
Once a labeled image has been generated, specific methods that rely on Deep Learning are used in order to actually create photo-realistic images that comply with the ASP-generated label.
It is worth noting that a careful implementation of proper ASP programs guarantees, by design, the generation of labels that comply with the domain knowledge at hand.
Furthermore, the use of ASP in this approach has some further advantages; indeed, one can also declaratively express specific desiderata that lead to the generation of images that can significantly differ from the original available ones also at a semantic level (e.g., number and position of elements, spatial relationships, etc.).

%it undergoes a refinement process based on deep learning techniques, for producing a synthetic yet realistic new image.
%
It is worth noting that image specifications go down to the pixel level; hence, given the combinatorial nature of the problem and the amount of involved information, ASP modeling must be carefully designed, in order to both correctly express knowledge/desiderata and reduce the computational cost.
% The herein presented approach makes use of existing labeled images for generating new ones via a set of ad-hoc defined ASP programs; such programs encodes the available medical knowledge and describe and model the desired structure of new images at the pixel level.
% Given the combinatorial nature of the problem and the amount of involved information, ASP modelling must be carefully designed, in order to both correctly express the medical knowledge and reduce the computational cost.
% Once a labeled image has been generated, it undergoes a refinement process based on deep learning techniques, for producing a synthetic yet realistic new image.

To the best of our knowledge, this is one of the first attempts of employing ASP in medical image generation and augmentation.
We tested our approach on the cited Laryngeal Endoscopic Dataset; results prove the viability of the approach, which is able to allow the generation of new images according to declaratively expressed directions.
%successfully generate realistic images in compliance with all the medical requirements.

\smallskip

In the following, we illustrate the proposed approach mainly focusing on the design and implementation of the ASP-based declarative generation of new labeled images.
The remainder of the paper is structured as follows. 
We first briefly introduce data augmentation techniques and their related work in Section~\ref{sec:data_aug}; in Section~\ref{sec:methods} we provide a detailed description of our approach, that has been tested in Section~\ref{sec:results_discussion}. 
We eventually draw our conclusions in Section~\ref{sec:conclusion}.

%%%%%%%%%%%%%%%%%%%%%%%%%%%%%%%%%%%%%%%%%%%%%%%%%%%%%%%%%%%%%%%%%%%
\section{Image Data Augmentation}\label{sec:data_aug}
Image data augmentation techniques have been widely studied in the literature and used in state-of-the-art solutions to reduce overfitting, increase generalizability and overcome the lack of data or other limitations that could affect algorithm performance. 
Indeed, data augmentation: $(i)$ is a method that results much less expensive than regular data collection with its label annotation, $(ii)$ can be extremely accurate (it is generated from ground-truth data), $(iii)$ controllable, to some extent, in generating balanced data~\cite{khalifa2022comprehensive}.

Traditionally, image data augmentation is performed relying on ``classical'' strategies or deep learning-based methods. In the first case, geometric transformation (i.e., flipping, rotation, shearing, cropping, translation in the geometric transformation) and photometric shifting (i.e., color space shifting, image filtering, addition of noise) are applied to existing available images in order to enrich the collection ~\cite{khalifa2022comprehensive}.
However, these techniques present some disadvantages, including memory consumption, transformation costs, and additional training time. 
Also, some photometric shifting strategies can produce the eliminations of important color information or specific features in the image, thus not always guaranteeing the preservation of nature and meaning of the image labels~\cite{shorten2019survey}.
On the other hand, DL methods, especially Generative Adversarial Networks (GAN)-based ones, represent a huge breakthrough in image generation, due to the ability to generate artificial images from the initial dataset and then make use of them to predict image features. 
GANs are composed of two networks: a {\em generator} network that creates tentative fake images and a {\em discriminator} network that identifies whether the generated images are indicative of real-world evidence or not ~\cite{aggarwal2021generative}.
Nevertheless, GANs are inherently unstable and suffer from both the lack of meaningful measures to evaluate the quality of their result and limited sample generation capabilities when only a little representative of the population is available.

In the biomedical context, the availability of huge datasets is one of the major concerns: it is indeed a difficult task, as it requires continuous efforts in the long term. 
Image data augmentation techniques aim at tackling this issue, generating medical images for automated assessment of pathological conditions, and supporting healthcare providers in finding the most appropriate preventive interventions and therapeutic strategies without the need for the availability of large medical datasets~\cite{chen2022generative}.

Kossen et al.~\cite{kossen2021synthesizing} used GANs to create synthetic brain data and corresponding labels, showing good performance in the arterial brain vessel segmentation task. 
Similarly, Toikkanen et al.~\cite{toikkanen2021resgan} used GAN to improve the quality of the predictive model in localizing the hemorrhage from computerized tomography (CT) scans. 
Synthetic samples from generative models have been demonstrated to alleviate the in-balance and scarcity of labeled training issues.
In the same context, Zhai et al.~\cite{zhai2022ass} proposed a novel asymmetric semi-supervised GAN (ASSGAN) to generate reliable segmentation-predicted masks. 
The authors show that in the absence of labeled data, the network can make use of unlabeled data to improve segmentation performance. 

To the best of our knowledge, there are no logic-based approaches for performing image data augmentation. 
Some attempts have been done of using ASP for improving the segmentation and the quality of medical images (e.g., ~\cite{bruno2021combining},~\cite{bruno2022dedudeep}); however  they don't focus on the  generation of new images, but rather are concerned in quality improvements.  
Other logic-based contributions concern the somehow related field of Content Generation, where Answer Set Programming has been used for the production of game contents with desirable properties~\cite{DBLP:conf/aiia/CalimeriGIPPT18,DBLP:books/daglib/0034882,DBLP:journals/tciaig/SmithM11}. 
Such approaches demonstrate that ASP can be used for declaratively express quantitative and qualitative desiderata as well as content generation strategies, which can be of use in the case of image generation, providing one with the possibility to easily increment, modify and update new knowledge at will.

%%%%%%%%%%%%%%%%%%%%%%%%%%%%%%%%%%%%%%%%%%%%%%%%%%%%%%%%%%%%%%%%%%%
\section{An ASP-based Method for Image Generation} \label{sec:methods}
In this section, we present our proposed ASP-based method for custom image generation. 
We focus on the case of laryngeal endoscopic image generation, and design and develop ad-hoc ASP programs for generating new synthetic images starting from the dataset in~\cite{laves2019dataset}. 
\begin{figure}
\centering
\subfigure{\includegraphics[width=0.5\textwidth]{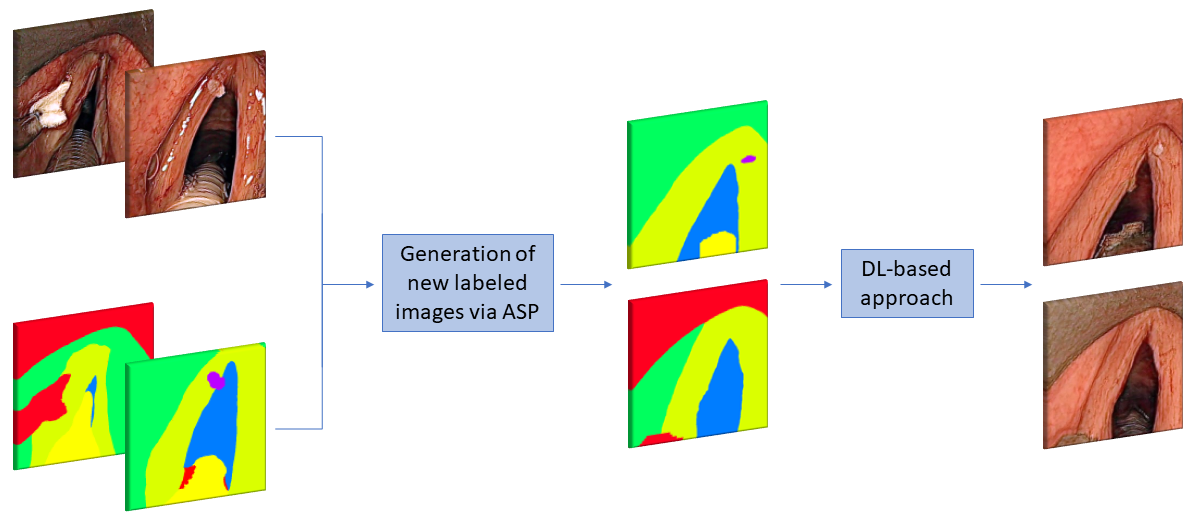}}
\caption{Workflow of the proposed approach}
\label{fig:worflow}
\end{figure}
Figure \ref{fig:worflow} shows the workflow of our approach: at first, a set of available photo-realistic images and their corresponding labels are used in order to identify (at a semantic level) what are the relevant elements that images can contain and that make them differ one from another, according to prior medical knowledge; then, a declarative module based on ASP is in charge to make use of such elements for generating new labeled images that comply to explicitly expressed desiderata. 
Eventually, once a labeled image is generated, a module based on DL methods is used to create photo-realistic images that match the ASP-generated directions.

In the following, we first recall the case study and then detail the framework.

%%%%%%%%%%%%%%%%%%%%%%%%%%%%%%%%%%%%%%%%%%%%%%%%%%%%%%%%%%%%%%%%%%%%%%%%%%%
\subsection{A Case Study: Laryngeal Endoscopic Images}\label{subsec:dataset}

\begin{figure}[h!]
    \begin{center}
    \subfigure[]{\includegraphics[width=0.17\textwidth]{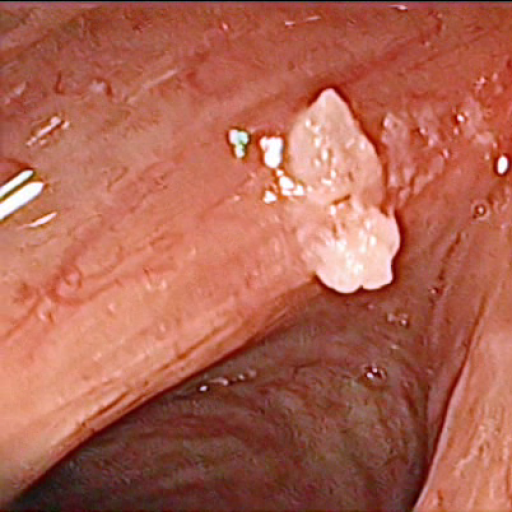}} 
    \subfigure[]{\includegraphics[width=0.17\textwidth]{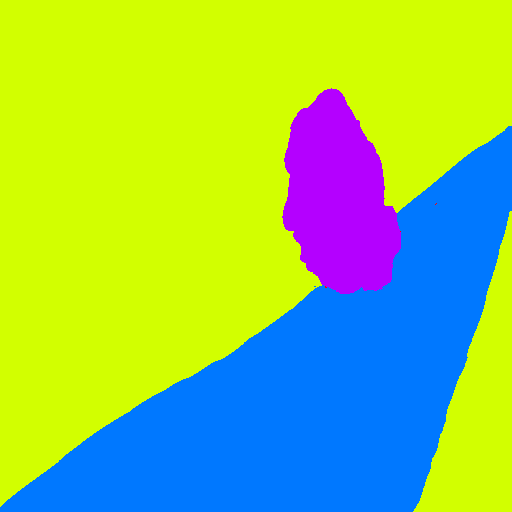}}

    \subfigure[]{\includegraphics[width=0.17\textwidth]{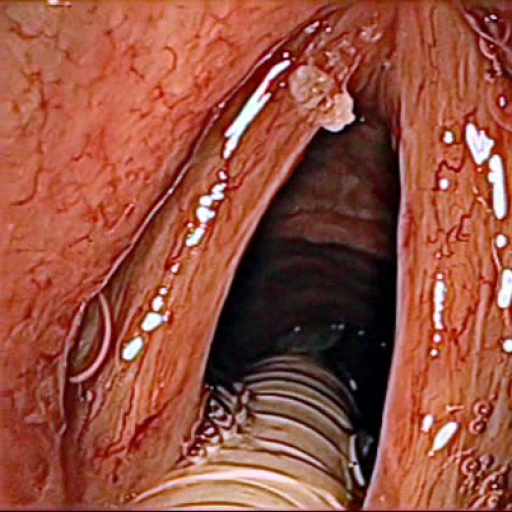}}
    \subfigure[]{\includegraphics[width=0.17\textwidth]{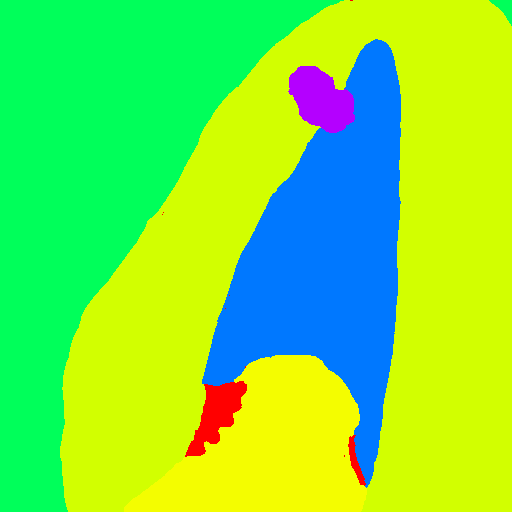}}

    \subfigure[]{\includegraphics[width=0.17\textwidth]{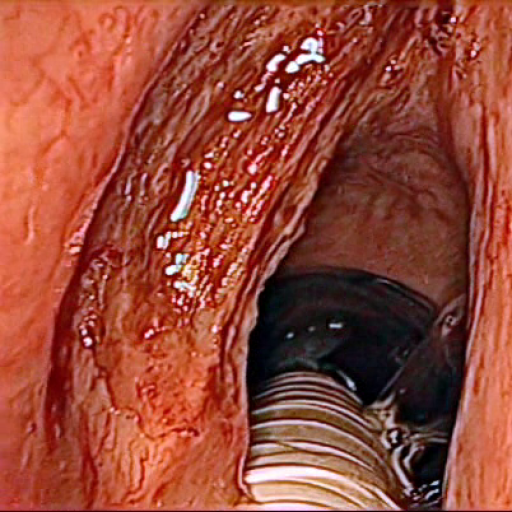}}
    \subfigure[]{\includegraphics[width=0.17\textwidth]{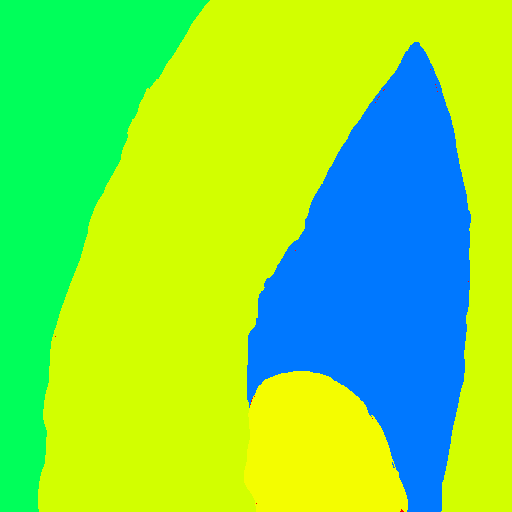}}    
    
    \subfigure[]{\includegraphics[width=0.17\textwidth]{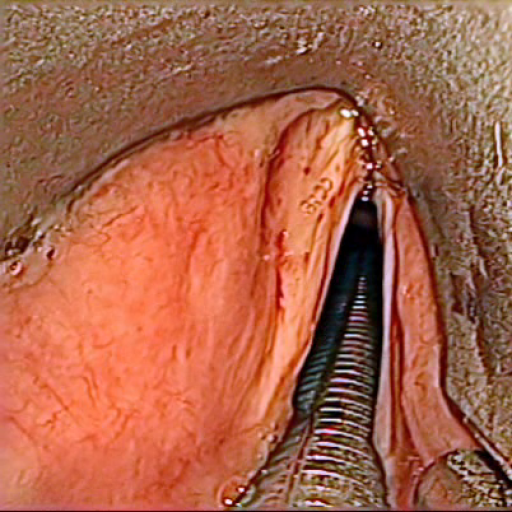}}
    \subfigure[]{\includegraphics[width=0.17\textwidth]{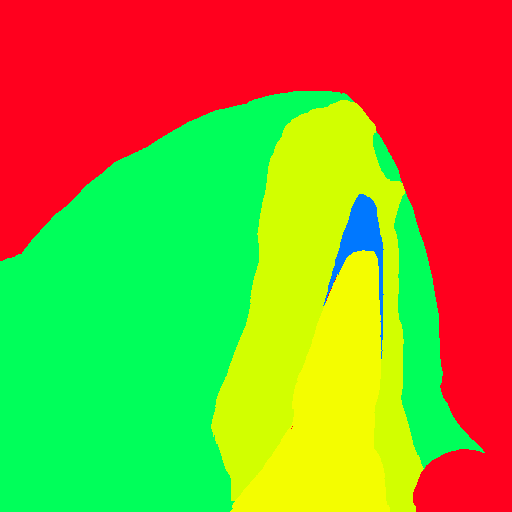}}

    \subfigure[]{\includegraphics[width=0.17\textwidth]{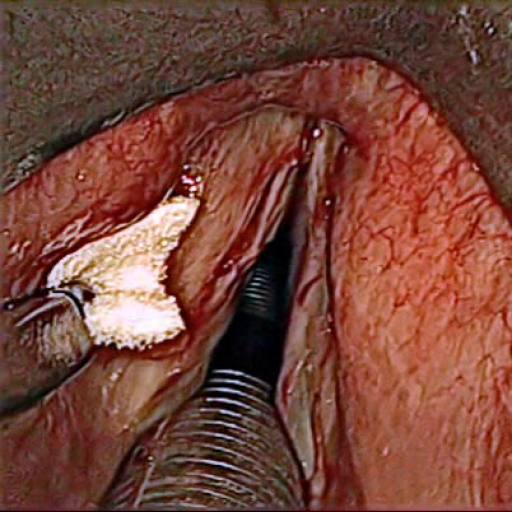}}
    \subfigure[]{\includegraphics[width=0.17\textwidth]{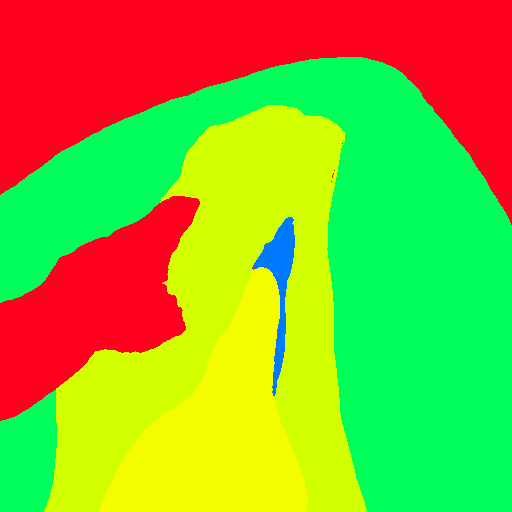}}
    \end{center}
    \caption{Examples of Laryngeal Endoscopic images from the dataset ~\cite{laves2019dataset}. Images come from Sequences 1,2,3,5, and 8 (from top to bottom). Raw images  and the corresponding labeled ones are reported in the first and second columns, respectively. 
    \label{fig:dataset_example}} 
\end{figure}
The Laryngeal Endoscopic Images dataset~\cite{laves2019dataset} consists of $536$ manually segmented in vivo color images ($512$×$512$ pixels) of the larynx captured from videos recorded during two different resection surgeries. 
The images are composed of $7$ classes: \void (in gray), \vocalfolds (in \colorVocalFolds), \othertissue (in \colorOtherTissue), \glottalspace (in \colorGlottalSpace), \pathology (in \colorPathology), \surgicaltool (in \colorSurgicalTool), and \intubation (in \colorIntubation).

The dataset features $8$ sequences from two patients; they have been categorized into $5$ different groups, depending on what kind of features the images exhibit.
This results in solid medical background knowledge, that is detailed, for each group, in the following (see~\cite{laves2019dataset}).

\begin{enumerate}
    \item Sequence $1$: these are images extracted from pre-operative videos; the tumor is always present: in particular, it is clearly visible on the vocal folds. 
    The images feature also changes in scale, translation, rotation, and do not present intubation, nor visible instruments.

    \item Sequence $2$: as in Sequence 1, images are extracted from pre-operative videos: the tumor is present and clearly visible; changes in scale and translation are featured. 
    In these sequences, however, there are visible instruments, with intubation.

    \item Sequences $3$--$4$: these images are extracted from post-operative videos; given that the tumor has been removed, there is no visible tumor. 
    Images feature changes in scale and translation, no visible instruments, yet intubation is present, along with some damaged tissue. 

    \item Sequences $5$--$7$: images taken from pre-operative videos, where  instruments are visible while manipulating and grasping the vocal folds. Changes in scale and translation, are featured; intubation is present.
 
    \item Sequence $8$: images extracted from post-operative videos. 
    Here, blood on vocal folds is visible, along with instruments,  surgical dressing, and intubation.
\end{enumerate}

A visual example of images in the dataset is represented in Fig.~\ref{fig:dataset_example}, where a raw image (top) is reported along with its corresponding labeled version (bottom) for each of the five groups.  
For instance, images (b),(g) come from Sequence $2$, where the tumor is clearly visible on the vocal folds and also intubation is present.

%%%%%%%%%%%%%%%%%%%%%%%%%%%%%%%%%%%%%%%%%%%%%%%%%%%%%%
%%%%%%%%%%%%%%%%%%%%%%%%%%%%%%%%%%%%%%%%%%%%%%%%%%%%%%%%%%%%%%%%%%%
\subsection{Framework Core}\label{subSec:generalApproach}
The proposed approach relies on an ASP-based model to generate new synthetic labeled images. 
Basically, logic programs are in charge of describing the main characteristics of a specific scenario and modeling the a-priori knowledge.
It is worth noting that the declarative nature of ASP allows us to $(i)$ incorporate explicit medical knowledge in the image generation process, $(ii)$ generate semantically labeled new images that comply with a set of provided requirements, $(iii)$ easily modify specific aspects of images according to variation of the domain or specific directions.
More in detail, declarative specifications not only can force realistic compliance in image generation, but can also allow to heavily customize the way new data will look like at a ``semantic'' level.
For instance, one can explicitly ask for images containing tumors of a given average size, or to include an additional instrument, or to prefer images featuring specific relative positions of some elements, and so on.

Intuitively, for each domain, ad-hoc ASP programs need to be properly designed; nonetheless, when working at a pixel level, several pieces of knowledge can be reused across different tasks and domains. 
In the following, we focus on the specific design choices we made for tackling the problem in the already mentioned laryngeal endoscopic domain.  

We started from a given number of labeled images taken from the dataset described in Section~\ref{subsec:dataset}, and identified the image elements to be considered as relevant in the data augmentation process; such elements are removed from the original labeled images and will be actually generated from scratch, while the others are just kept. 
More specifically, we decided to keep the classes \vocalfolds, \glottalspace  and \othertissue as the background of the images; indeed, their shape and appearance do not significantly vary between different images.
On the contrary, we focused on the classes \intubation and \surgicaltool for their generation from scratch, as during the surgery they move faster than background tissues and, consequently, their shape and position can easily change according to the rotation and the angle of endoscopic video images. Furthermore, we also considered \pathology class: its importance trivially derives from the domain, and its occurrence in the whole dataset is significantly lower ~\cite{laves2019dataset}; the generation of further images featuring the \pathology class allows to better balance the dataset.

% For the generation of the aforementioned classes, we have taken into account two different strategies. \%todo[inline]{aggiustare e completare da qui in poi}
% %
% Specifically, the first one starts from a selection \%todo[inline]{dire una chiacchiera su come sono selezionate queste segmentazioni} of different segmentations of the chosen class; then a segmentation is guessed along with a pivot, a point into a suitable part of the image where the selected segmentation has to be placed.
% %
% Instead, the second strategy relies on the generation from scratch of the chosen class, starting from a central point from which construct and design, step by step, the rest of the class under consideration.
%
%
% ... perchè abbiamo scelto il secondo approccio?\%todo[inline]{aaaah}\%todo[inline]{dire che lavorare a dimensioni tipo scatole cinesi e' cruciale per motivi di performance ed e' cosa che si fa piuttosto spesso (si veda \cite{DBLP:conf/aiia/CalimeriGIPPT18})}
%\%todo[inline]{a simona non convince il fatto di menzionare i due approcci e poi sviluppare solo il secondo, perciò abbiamo evitato di parlare del primo. Kali, che ne pensi?}

For generating and placing objects as instances of the aforementioned classes, we rely on the following strategy: given a class, a point is properly selected that is intended to represent the position of the object to construct (as its ``center''); starting from the chosen pixel we design and construct, step by step, the remaining part of the object under consideration.

As presented in Section~\ref{subsec:dataset}, the images in the dataset are collected in 5 different groups: images in each group have specific characteristics that differ among groups, and are related to the classes of objects that can be featured in images and their relative positions. We want new images to satisfy such features, accordingly.
%
%We have to model the medical knowledge  separately for each group.
%However, given that the background is almost the same across groups, we decided to 
To this aim, we designed ad-hoc ASP programs for the generation of different kind of objects, namely in the \pathology, \intubation, and \surgicaltool classes. 
labeled images for a given group are obtained by properly combining the results from some of such programs (as an example, we can create images of the group $2$ by generating a \pathology object and the \intubation).
In the following, we describe the ASP programs along with input and output information.
We assume that the reader is familiar with standard ASP syntax and semantics; for more details, we refer to the vast literature (e.g.,~\cite{DBLP:journals/tplp/CalimeriFGIKKLM20}).

\paragraph{Input.}\ The ASP programs take in input $512 \times 512$ bitmap images, in form of matrices of the same size and properly represented as facts.
Each matrix element is associated with a color determined by the class of the object present in that position. 
In particular, as said above, we consider images where \vocalfolds, \glottalspace and \othertissue classes are fixed; thus, the input matrices contain elements (or cells) already colored in \colorVocalFolds, \colorGlottalSpace and \colorOtherTissue.
In order to properly address the resulting search space, we adopt an approach inspired by what in Content Generation contexts is called ``space partitioning''~\cite{DBLP:conf/fdg/TogeliusKSY11,DBLP:conf/aiia/CalimeriGIPPT18}, that works by dividing large areas into smaller zones to be addressed separately; the final result is then obtained by combining the partial results.
In particular, we iterate through the elements of the matrix by considering blocks of dimension $64 \times 64$; each block is, in turn, divided into sub-blocks of dimension $8 \times 8$. Such dimensions are empirically chosen.

\noindent Matrix elements are modeled by 
 facts of the form
%\begin{center}
    \verb|cell(X,Y,Col,IDB,IDSB)|,
%\end{center}
%
\noindent where term variables \verb|X|, \verb|Y|, \verb|Col|, \verb|IDB|, and \verb|IDSB| are mapped to, respectively, the rows and columns of the matrix, the color associated to that cell, the identifier of the block and the identifier of the sub-block containing the cell \verb|(X,Y)|.

\paragraph{Output.}\ Given the class of the object to generate, the ASP program identifies a suitable area of the matrix where the object can be placed.
This is done by generating predicates of the form \verb|subBlockIn(IDB,IDSB)|, that represent the sub-blocks \verb|IDSB| within the block \verb|IDB|, whose cells will be re-colored according to the color of the class of interest.

\paragraph{ASP program for generating \pathology objects.} From the medical background, we know that in each image where the \pathology is present (images in groups $1$ and $2$), this is clearly visible and positioned ``on top of'' the class \vocalfolds. 
Thus, the aim of the ASP program is to get as input a matrix where the \vocalfolds are represented via cells colored in \colorVocalFolds, and generate a new tumor by properly select a number of sub-blocks to be re-colored in \colorPathology.
The first part of the ASP program consists of a series of guessing rules aiming at choosing: $(i)$ the block (and therefore the area) where the tumor is placed, $(ii)$ the position (i.e., the ``center'' point) in such block, and $(iii)$ some {\em contour ``pivot'' points} that will be used to define final shape and size of the tumor; we experimentally determined that $8$ pivot points are suitable for our purposes.
To encode $(i)$, we define the choice rule
\begin{footnotesize}
\begin{verbatim}
{chosenBlock(ID) : lightGreenBlock(ID)} = 1.
\end{verbatim}    
\end{footnotesize}

\noindent where atom \verb|lightGreenBlock(ID)| represents all the suitable blocks on which it is possible to generate the tumor.
It is worth noting that this ensures that the tumor will only appear where it is supposed to be according to the background knowledge.
Whereupon, based on from \verb|chosenBlock(ID)|, we encode $(ii)$ via the choice rule
\begin{footnotesize}
 \begin{verbatim}
{center(ID,X,Y) : centralPoint(ID,X,Y)} = 1 
:- chosenBlock(ID).
\end{verbatim}
\end{footnotesize}

\noindent where \verb|centralPoint(ID,X,Y)| collects all the possible starting points, i.e., points previously selected that are far enough from the area of the matrix that do no comply with the presence of the tumor, and hence are suitable for its construction.
Eventually, to encode $(iii)$, we draw the diagonals with respect \verb|center(ID,X,Y)| and \verb|chosenBlock(ID)|, and guess eight distinct points positioned as follows: two on the main diagonal (predicate \verb|sameMainDiag|), two on the secondary diagonal (\verb|sameSecDiag|), two on the same row with respect to the central point (\verb|sameRow|), and two on the same column with respect to the central point (\verb|sameCol|).
This is encoded by the following choice rules: \\

\begin{footnotesize}
\begin{verbatim}
{pointsMainDiag(ID,X,Y) : sameMainDiag(ID,X,Y)} = 2 
:- chosenBlock(ID).

{pointsSecDiag(ID,X,Y) : sameSecDiag(ID,X,Y)} = 2 
:- chosenBlock(ID).

{horizPoints(ID,X,Y) : sameRow(ID,X,Y)} = 2
:- chosenBlock(ID).

{vertPoints(ID,X,Y) : sameCol(ID,X,Y)} = 2 
:- chosenBlock(ID).
\end{verbatim}
\end{footnotesize}

Furthermore, we want to ensure that the guessed points satisfy some geometric properties, define an ordering among them depending on their position, and assign them with an identifier, accordingly. 
In such a way, we obtain the $8$ contour pivot points that are represented via  atoms \verb|contourPivot(IDFP,ID,X,Y)|,
where \verb|IDFP| is the identifier of the contour pivot and \verb|ID| is the identifier of the block containing the cell \verb|(X,Y)| of the guessed contour pivot.
Moreover, we make use of some constraints to avoid situations in which the guessed contour pivot points do not comply with a realistic situation (for instance, a contour pivot point is placed too close to the central point).
On the basis of the suitable contour pivot points, we define the outline of the tumor: for each consecutive pair of pivot points, we define a connection by guessing a path over the sub-blocks of the \verb|chosenBlock|. 
In particular, given two contour pivot points \verb|contourPivot(IDFP1,ID,X1,Y1)| and
\verb|contourPivot(IDFP2,ID,X2,Y2)|, connecting paths are chosen within a restricted area of the matrix that corresponds to the rectangle having vertices $(X1,Y1), (X1,Y2), (X2,Y1)$ and $(X2,Y2)$. 
Relying on such sort of ``bounding boxes'' has two different purposes. 
First, we reduce the search space: indeed, guessing over the whole (bigger) area would be prohibitive, given the combinatorial nature of the problem;
furthermore, we model what intuitively would be done by a human expert on the basis of the knowledge about the realistic shape of a tumor (for instance, one should avoid that borders feature significant protrusions or cavities).
The size of the rectangular area can be adapted, and experimentally set to values representing good trade-off between the number of sub-blocks to be guessed and the number and the shape of paths to be explored.

Sub-blocks inside the defined rectangular area are considered as ``guessable", a concept modeled by instances of the predicate  
\verb|guessableSubBlock(IDB,IDSB)|,
where \verb|IDSB| identifies a sub-block inside the block \verb|IDB| that is suitable to be part of the path.
The path between two given contour pivot points is defined via the following rules:

\begin{footnotesize}
\begin{verbatim}
subBlockIn(ID,IDS) | subBlockout(ID,IDS) 
:- guessableSubBlock(ID,IDS).

subBlockIn(ID,IDS) 
:- contourPivot(_,ID,X,Y), cell(X,Y,_,ID,IDS). 

reachSubBlock(ID1,IDS1,ID2,IDS2) 
:- subBlockIn(ID1,IDS1), subBlockIn(ID2,IDS2), 
IDS1!=IDS2, adjSubBlock(ID1,IDS1,ID2,IDS2).

reachSubBlock(ID1,IDS1,ID3,IDS3) 
:- reachSubBlock(ID1,IDS1,ID2,IDS2), 
reachSubBlock(ID2,IDS2,ID3,IDS3),IDS1!=IDS3.
\end{verbatim}
\end{footnotesize}

Briefly, the rules in the program snippet above guess which sub-blocks can be part of the path, enforce that the contour pivot points are part of the path, and check reachability via recursion. 

We express our preference among possible paths via proper weak-constraints. 
For instance, the following one has been defined to state that we prefer paths where non-adjacent sub-blocks are not in-line (thus trying to avoid both {\em zigzags} and straight lines).

\begin{footnotesize}
\begin{verbatim}
:~ subBlockIn(ID,IDS), subBlockIn(ID1,IDS1), ID!=ID1,
cell(X,Y,_,ID,IDS), cell(X,Y1,_,ID1,IDS1),
not adjSubBlock(ID,IDS,ID1,IDS1).  [1@1, IDS,IDS1] 
\end{verbatim}
\end{footnotesize}

\paragraph{ASP program for generating \intubation.}\ Images featuring \intubation are from groups $2$--$5$. 
In these images, \intubation is present in the bottom part of the photo, starting from the border; furthermore, it is always positioned on top of the \glottalspace class. 
Thus, the ASP program is designed in order to choose a proper set of sub-blocks belonging to the \glottalspace, hence colored in \colorGlottalSpace, where the \intubation has to be positioned; such sub-blocks are recolored as \intubation, i.e. \colorIntubation. 

Sub-blocks are selected by a program that behaves similarly to the one designed for the \pathology case; 
the main differences are in the way guessable sub-blocks and contour pivot points are identified. 
In particular, since the intubation has to be positioned on the glottal space, the guessable sub-blocks are selected in the \colorGlottalSpace area; 
moreover, the center point is chosen within the very lowest region of the matrix, and all contour pivot points are guessed above the center point, such that, by connecting them, a shape resembling a semi-oval is obtained (as this is the form an expert expect to see when finding the intubation in a picture). 
Cells within this shape are colored in \colorIntubation.

\paragraph{ASP program for generating \surgicaltool.}\
The ASP program for generating  \surgicaltool objects is defined according a strategy similar to what described above. 
Differences are mainly related to the way shapes of surgical tools are modeled (this clearly depends on the type of tool) and their positions are chosen (as tools can be present both on top of the \vocalfolds and in the \glottalspace).

%%%%%%%%%%%%%%%%%%%%%%%%%%%%%%%%%%%%%%%%%%%%%%%%%%%%%%%%%%%%%%%%%%%%%%%%%%%
\subsection{From Labels to Raw Data}  \label{subsec:dl}
The labeled images generated via the ASP-based method described above represent semantic descriptions of images that comply with the background knowledge and the expressed desiderata. 
The next step is to generate a photo-realistic counterpart for each one: intuitively, new raw images are supposed to be such that, when semantically segmented, correspond to the related output of the ASP-based phase.

Among all different types of image synthesis tasks, {\em label-to-image} is one of the most challenging ones, due to the complexity of the images that have to be synthesized~\cite{zhu2022label}. 
Different works have been proposed in the literature, some for addressing paired-data training (i.e., the model is fed with label maps and corresponding images), and others for unpaired-data (i.e., unpaired label maps and images are used for training)~\cite{zhu2022label}. 
In the scope of this work, we are experimenting with one of the most recent state-of-the-art proposals: Semantic Image Synthesis With Spatially-Adaptive Normalization
(SPADE)~\cite{park2019semantic} that is commonly used as paired-data techniques. % as a possible solution for our label-to-image translation. 
SPADE processes the input semantic layout (i.e., an abstract representation of an image that defines the different parts or objects in the image and their spatial relationships) through several layers of convolution, normalization, and nonlinearity. Instead of traditional normalization layers, the authors used spatially-adaptive normalization layers. These layers modulate the activations using the input semantic layout through a learned transformation that adapts to the spatial layout. This helps the network to effectively propagate semantic information and produce better results than previous methods \cite{park2019semantic}.
Specifically, in the SPADE approach, the mask is first projected onto an
embedding space and then convolved to produce the modulation parameters. Unlike prior conditional normalization methods, these parameters are not vectors, but tensors with
spatial dimensions that are multiplied
and added to the normalized activation element-wise. 
In our experimental analysis, we used the same parameters configuration provided by the authors (i.e., learning rates of 0.0001 for the generator and 0.0004 for the discriminator and ADAM as optimizer) and we trained the network for 350 epochs.

%%%%%%%%%%%%%%%%%%%%%%%%%%%%%%%%%%%%%%%%%%%%%%%%%%%%%%%%%%%%%%%%%%%
\section{Tests and Results}\label{sec:results_discussion}
In this section, we present the results obtained in the first tests of the proposed framework. 
In particular, we show how a new labeled image is produced by generating the tumor, the intubation, and, eventually, surgical tools, by providing visual examples of the results.

\begin{figure}[h!]
    \begin{center}
    
    \subfigure[]{\includegraphics[width=0.17\textwidth]{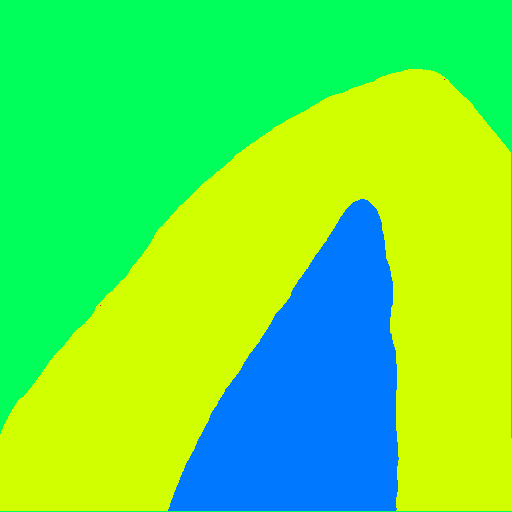}}
    
    \subfigure[]{\includegraphics[width=0.17\textwidth]{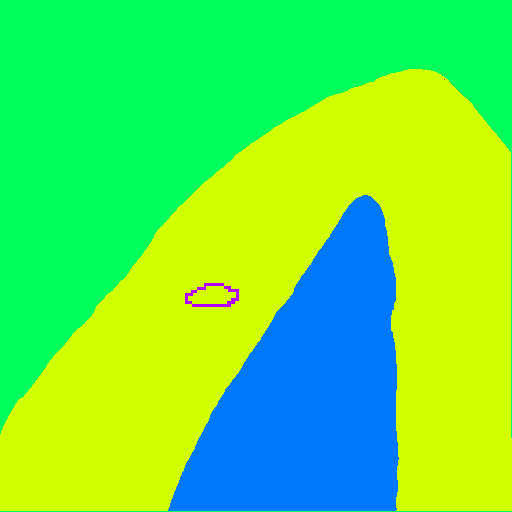}}
    \subfigure[]{\includegraphics[width=0.17\textwidth]{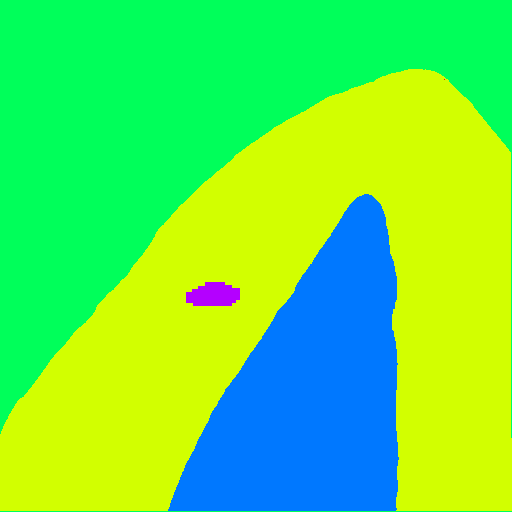}}

    \subfigure[]{\includegraphics[width=0.17\textwidth]{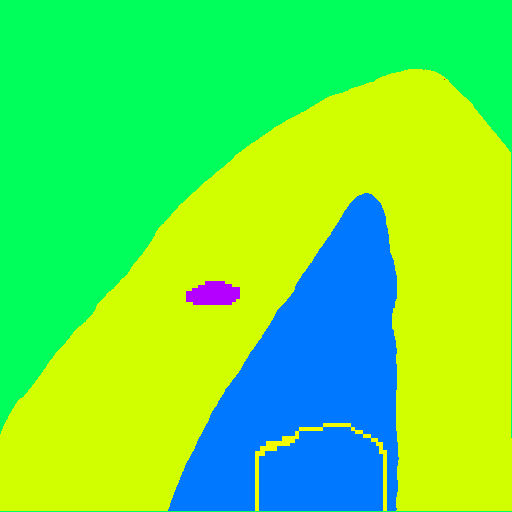}}
    \subfigure[]{\includegraphics[width=0.17\textwidth]{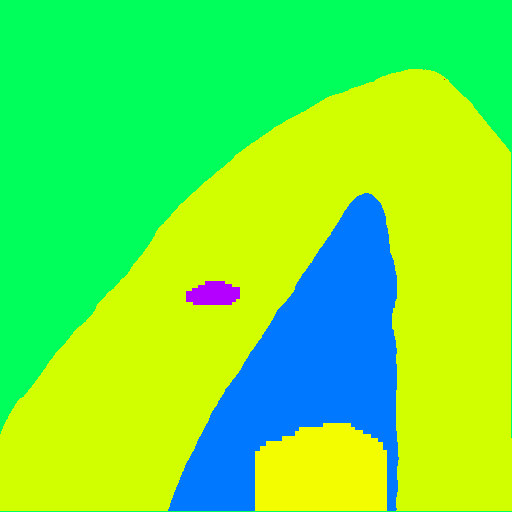}}
    
    \subfigure[]{\includegraphics[width=0.17\textwidth]{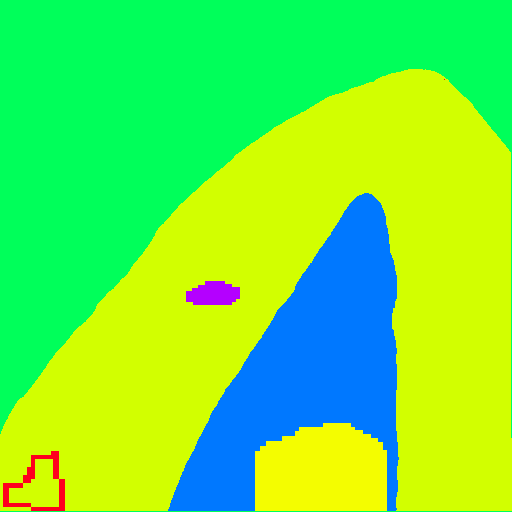}}
    \subfigure[]{\includegraphics[width=0.17\textwidth]{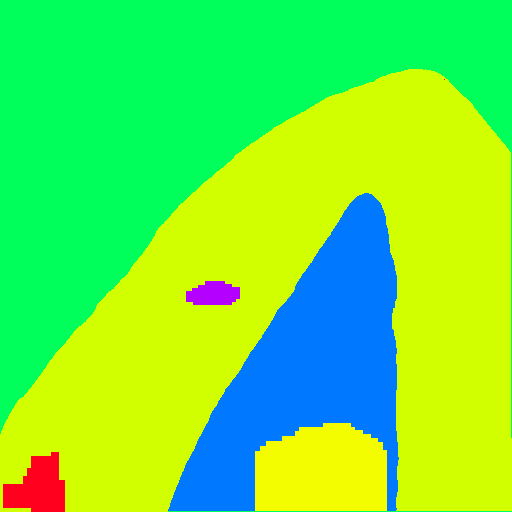}}    
    \end{center}
    
    \caption{Generation from scratch of classes \pathology, \intubation, and \surgicaltool.}\label{fig:runningExample} 
\end{figure}

As discussed in Section~\ref{subSec:generalApproach}, the ASP programs are fed with input images featuring only the classes \othertissue, \vocalfolds, and \glottalspace, colored in \colorOtherTissue, \colorVocalFolds, and \colorGlottalSpace, respectively; an example is reported in see Figure~\ref{fig:runningExample}(a).

For generating a sample labeled image of group $1$, and in particular Sequence 1, we can start from this image and generate a tumor over it; such sequence images, indeed, do not feature intubation nor surgical tools.
%

%%%%%%%%%%%%%%%%%%%%%%%%%%%%%%%

By taking advantage from the ASP programs described in Section~\ref{subSec:generalApproach}, a center point and then series of contour pivot points are generated and then properly connected, thus forming the shape of a tumor over the correct area (as reported in the dataset description, the tumor must be clearly visible on the vocal folds, i.e., \colorVocalFolds area): see Figure~\ref{fig:runningExample}(b).
Eventually, we assign the appropriate color (i.e., \colorPathology) to each pixel inside the obtained area; the final result is shown in Figure~\ref{fig:runningExample}(c).

In order to generate images of Groups $2$--$5$, we need to incorporate the intubation class (see Figure~\ref{fig:runningExample}(d) and~\ref{fig:runningExample}(e)) and the surgical tool (see Figure~\ref{fig:runningExample}(f) and~\ref{fig:runningExample}(g)).
For instance, note that an object of intubation class, according to the a-priori knowledge and in order to generate a realistic result, must be positioned and the contour must be guessed inside the \glottalspace area (\colorGlottalSpace).

It is worth noting that different answer sets model different positions, sizes, and shapes;
what is reported in Figure~\ref{fig:runningExample} shows the result of randomly chosen ones.
We also point out that by slightly modifying the ASP programs one can obtain significantly different  results; for instance, one might change the number of contour pivot points, the way are placed and distanced, how they are connected in order to form the shape, and so on.

The preliminary tests we carried out show that the herein proposed approach is viable, even under some {\em caveats}.
Indeed, for the whole process to be successful, significant efforts must be spent on outlining and formally representing the background knowledge, for properly designing the ASP programs and fine-tuning them in order to make the newly generated data match the original desiderata, and for defining the workflow for composing the result. 
Furthermore, the computational burden to be carried out by ASP solvers is heavy, especially in the case of image generation carried out at the pixel level.
However, the advantages of declarative specifications that actually guide the generation of new data clearly overcome such considerations, especially if one considers that, in a typical scenario, data are generated once and used many.

\begin{figure}

    \begin{center}
    \subfigure[]{\includegraphics[width=0.4\textwidth]{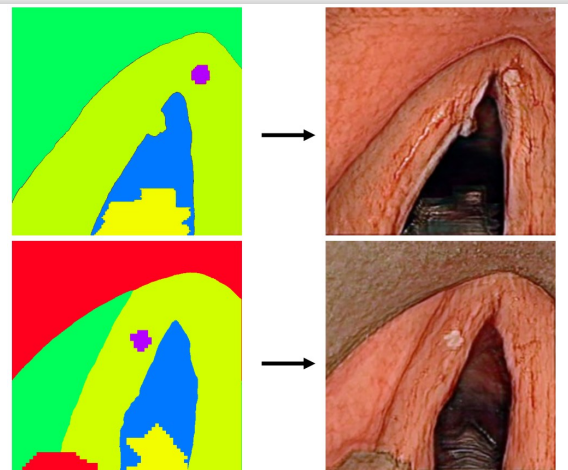}}
    
    \caption{Example of synthetic image generation from ASP-based labels.}
    \label{fig:label_to_raw}
    \end{center}
\end{figure}

Starting from the labeled image generated via ASP, we make use of SPADE (See Section \ref{subsec:dl}) to create photo-realistic data. As shown in Figures \ref{fig:label_to_raw}, our approach is able to successfully generate synthetic images from semantic labels.

Some additional considerations are reported in the next Section.

%%%%%%%%%%%%%%%%%%%%%%%%%%%%%%%%%%%%%%%%%%%%%%%%%%%%%%%%%%%%%%%%%%%
\section{Conclusion and Perspectives}
\label{sec:conclusion}
In this paper we presented a framework aiming at enabling the declarative specification of data augmentation processes; in particular, we proposed the use of Answer Set Programming for guiding the generation of realistic images in the biomedical domain. 
The presented approach relies on the collection of a (small) dataset of labeled images from the ground truth, and on the identification of relevant elements that make images differ one from another; then, specific ASP reasoning tasks are employed for generating brand new labeled images, obtained by describing how such elements can appear in the images and then properly composing the output.
The new semantically labeled images are then used as input for specific methods relying on Deep Learning for producing photo-realistic images, which actually constitute the final output. 
We assessed the viability of the approach over images coming from laryngeal endoscopic surgery videos, and the results are promising, as they show that declarative specifications can be incorporated in the image data augmentation process.
Such specifications, expressed via ASP, can encode both background knowledge and specific desiderata; in our opinion, this is one of the main strength points of the approach.
Indeed, it allows to significantly customize the generation of new raw data in a declarative fashion without the need for finding, collecting, and adapting data in the domain at hand (for instance, surgical images featuring a given number of instruments, a specific position of the tumor, etc.); and yet, it allows to enjoy the typical resilience of ASP with respect to knowledge update (i.e., changes in specifications).
As an example, one can easily adapt the logic programs so to change the number of elements of a given class, or the spatial relationships among elements (e.g., generating images featuring smaller/larger tumors or less/more instruments, etc.).

From a larger perspective, the use of ASP (as the declarative formalism of choice) in the loop of data augmentation to the extent herein described allows one to collect declarative specifications (i.e., logic programs) and ``translate'' them in such a way (i.e., properly generated labeled images) that they can be fed to DL methods. 
In particular, SPADE achieved satisfactory results, generating realistic raw images complying with a corresponding labeled one.
As future work is concerned, next steps involve experimental campaigns designed for assessing the quality of images with respect to the desired task and the performance of our approach on additional biomedical datasets.

%%%%%%%%%%%%%%%%%%%%%%%%%%%%%%%%%%%%%%%%%%%%%%%%%%%%%%%%%%%%%%%%%%%%%%%%%%%

\bibliography{main}

\begin{thebibliography}{10}

\bibitem{aggarwal2021generative}
Alankrita Aggarwal, Mamta Mittal, and Gopi Battineni, `Generative adversarial
  network: An overview of theory and applications', {\em International Journal
  of Information Management Data Insights}, {\bf 1}(1),  100004, (2021).

\bibitem{bruno2022dedudeep}
Pierangela Bruno, Francesco Calimeri, and Cinzia Marte, `Dedudeep: An
  extensible framework for combining deep learning and asp-based models', in
  {\em Logic Programming and Nonmonotonic Reasoning: 16th International
  Conference, LPNMR 2022, Genova, Italy, September 5--9, 2022, Proceedings},
  pp. 505--510. Springer, (2022).

\bibitem{bruno2021combining}
Pierangela Bruno, Francesco Calimeri, Cinzia Marte, and Marco Manna, `Combining
  deep learning and asp-based models for the semantic segmentation of medical
  images', in {\em Rules and Reasoning: 5th International Joint Conference,
  RuleML+ RR 2021, Leuven, Belgium, September 13--15, 2021, Proceedings 5}, pp.
  95--110. Springer, (2021).

\bibitem{DBLP:journals/tplp/CalimeriFGIKKLM20}
Francesco Calimeri, Wolfgang Faber, Martin Gebser, Giovambattista Ianni, Roland
  Kaminski, Thomas Krennwallner, Nicola Leone, Marco Maratea, Francesco Ricca,
  and Torsten Schaub, `Asp-core-2 input language format', {\em Theory Pract.
  Log. Program.}, {\bf 20}(2),  294--309, (2020).

\bibitem{DBLP:conf/aiia/CalimeriGIPPT18}
Francesco Calimeri, Stefano Germano, Giovambattista Ianni, Francesco Pacenza,
  Armando Pezzimenti, and Andrea Tucci, `Answer set programming for declarative
  content specification: {A} scalable partitioning-based approach', in {\em
  AI*IA 2018 - Advances in Artificial Intelligence - XVIIth International
  Conference of the Italian Association for Artificial Intelligence, Trento,
  Italy, November 20-23, 2018, Proceedings}, eds., Chiara Ghidini, Bernardo
  Magnini, Andrea Passerini, and Paolo Traverso, volume 11298 of {\em Lecture
  Notes in Computer Science}, pp. 225--237. Springer, (2018).

\bibitem{chen2022generative}
Yizhou Chen, Xu-Hua Yang, Zihan Wei, Ali~Asghar Heidari, Nenggan Zheng,
  Zhicheng Li, Huiling Chen, Haigen Hu, Qianwei Zhou, and Qiu Guan, `Generative
  adversarial networks in medical image augmentation: a review', {\em Computers
  in Biology and Medicine},  105382, (2022).

\bibitem{durgadevi2021generative}
M~Durgadevi et~al., `Generative adversarial network (gan): a general review on
  different variants of gan and applications', in {\em 2021 6th International
  Conference on Communication and Electronics Systems (ICCES)}, pp. 1--8. IEEE,
  (2021).

\bibitem{khalifa2022comprehensive}
Nour~Eldeen Khalifa, Mohamed Loey, and Seyedali Mirjalili, `A comprehensive
  survey of recent trends in deep learning for digital images augmentation',
  {\em Artificial Intelligence Review},  1--27, (2022).

\bibitem{kossen2021synthesizing}
Tabea Kossen, Pooja Subramaniam, Vince~I Madai, Anja Hennemuth, Kristian
  Hildebrand, Adam Hilbert, Jan Sobesky, Michelle Livne, Ivana Galinovic,
  Ahmed~A Khalil, et~al., `Synthesizing anonymized and labeled tof-mra patches
  for brain vessel segmentation using generative adversarial networks', {\em
  Computers in biology and medicine}, {\bf 131},  104254, (2021).

\bibitem{laves2019dataset}
Max-Heinrich Laves, Jens Bicker, L{\"u}der~A Kahrs, and Tobias Ortmaier, `A
  dataset of laryngeal endoscopic images with comparative study on convolution
  neural network-based semantic segmentation', {\em International journal of
  computer assisted radiology and surgery}, {\bf 14}(3),  483--492, (2019).

\bibitem{park2019semantic}
Taesung Park, Ming-Yu Liu, Ting-Chun Wang, and Jun-Yan Zhu, `Semantic image
  synthesis with spatially-adaptive normalization', in {\em Proceedings of the
  IEEE/CVF conference on computer vision and pattern recognition}, pp.
  2337--2346, (2019).

\bibitem{DBLP:books/daglib/0034882}
Noor Shaker, Julian Togelius, and Mark~J. Nelson, {\em Procedural Content
  Generation in Games}, Computational Synthesis and Creative Systems, Springer,
  2016.

\bibitem{shorten2019survey}
Connor Shorten and Taghi~M Khoshgoftaar, `A survey on image data augmentation
  for deep learning', {\em Journal of big data}, {\bf 6}(1),  1--48, (2019).

\bibitem{DBLP:journals/tciaig/SmithM11}
Adam~M. Smith and Michael Mateas, `Answer set programming for procedural
  content generation: {A} design space approach', {\em {IEEE} Trans. Comput.
  Intell. {AI} Games}, {\bf 3}(3),  187--200, (2011).

\bibitem{DBLP:conf/fdg/TogeliusKSY11}
Julian Togelius, Emil Kastbjerg, David~C. Schedl, and Georgios~N. Yannakakis,
  `What is procedural content generation?: Mario on the borderline', in {\em
  Proceedings of the 2nd International Workshop on Procedural Content
  Generation in Games, PCGames '11, Bordeaux, France, June 28, 2011}, pp.
  3:1--3:6. {ACM}, (2011).

\bibitem{toikkanen2021resgan}
Miika Toikkanen, Doyoung Kwon, and Minho Lee, `Resgan: Intracranial hemorrhage
  segmentation with residuals of synthetic brain ct scans', in {\em Medical
  Image Computing and Computer Assisted Intervention--MICCAI 2021: 24th
  International Conference, Strasbourg, France, September 27--October 1, 2021,
  Proceedings, Part I 24}, pp. 400--409. Springer, (2021).

\bibitem{wu2019review}
Hao Wu, Qi~Liu, and Xiaodong Liu, `A review on deep learning approaches to
  image classification and object segmentation', {\em Computers, Materials \&
  Continua}, {\bf 60}(2), (2019).

\bibitem{zhai2022ass}
Donghai Zhai, Bijie Hu, Xun Gong, Haipeng Zou, and Jun Luo, `Ass-gan:
  Asymmetric semi-supervised gan for breast ultrasound image segmentation',
  {\em Neurocomputing}, {\bf 493},  204--216, (2022).

\bibitem{zhu2022label}
Junchen Zhu, Lianli Gao, Jingkuan Song, Yuan-Fang Li, Feng Zheng, Xuelong Li,
  and Heng~Tao Shen, `Label-guided generative adversarial network for realistic
  image synthesis', {\em IEEE Transactions on Pattern Analysis and Machine
  Intelligence}, (2022).

\end{thebibliography}
\end{document}